\documentclass{article}
\usepackage{spconf,amsmath,epsfig}
\usepackage{graphicx}
\usepackage{amsmath}
\usepackage{amssymb}
\usepackage{amsthm}
\usepackage{booktabs}
\usepackage{algorithm}
\usepackage{algorithmic}
\usepackage{appendix}
\usepackage{newfloat}
\usepackage{listings}
\usepackage[dvipsnames, svgnames, x11names]{xcolor}
\usepackage{multirow}
\usepackage[switch]{lineno}

\let\OLDthebibliography\thebibliography
\renewcommand\thebibliography[1]{
  \OLDthebibliography{#1}
  \setlength{\parskip}{0pt}
  \setlength{\itemsep}{0pt plus 0.3ex}
}

\begin{document}\sloppy

\def\x{{\mathbf x}}
\def\L{{\cal L}}

\title{Weakly-Supervised Text Instance Segmentation}
%
\name{Xinyan Zu, Haiyang Yu, Bin Li, and Xiangyang Xue}
\address{\{xyzu20, hyyu20, libin, xyxue\}@fudan.edu.cn\\
Shanghai Key Laboratory of Intelligent Information Processing\\
School of Computer Science, Fudan University}

\maketitle

\begin{abstract}
Text segmentation is a challenging vision task with many downstream applications. Current text segmentation methods require pixel-level annotations, which are expensive in the cost of human labor and limited in application scenarios. In this paper, we take the first attempt to perform weakly-supervised text instance segmentation by bridging text recognition and text segmentation. The insight is that text recognition methods provide precise attention position of each text instance, and the attention location can feed to both a text adaptive refinement head (TAR) and a text segmentation head. Specifically, the proposed TAR generates pseudo labels by performing two-stage iterative refinement operations on the attention location to fit the accurate boundaries of the corresponding text instance. Meanwhile, the text segmentation head takes the rough attention location to predict segmentation masks which are supervised by the aforementioned pseudo labels. In addition, we design a mask-augmented contrastive learning by treating our segmentation result as an augmented version of the input text image, thus improving the visual representation and further enhancing the performance of both recognition and segmentation. The experimental results demonstrate that the proposed method significantly outperforms weakly-supervised instance segmentation methods on ICDAR13-FST (18.95$\%$ improvement) and TextSeg (17.80$\%$ improvement) benchmarks.
\end{abstract}
\section{Introduction}
In recent years, the text segmentation task has aroused growing research interests because it plays an important role in computer vision research and many practical applications, \textit{e.g.}, font style transfer, scene text removal, and interactive text image editing. Therefore, the text segmentation task can facilitate these downstream tasks and alleviate massive human labor for manually labeling pixel-level text regions.

The early work on text segmentation mainly follows the design of semantic segmentation for generic objects, which uses pixel-level annotations as supervision. Although the existing text segmentation methods have achieved considerable performance on several benchmarks, these methods rely heavily on a large amount of data with pixel-level annotations, \textit{e.g.}, the method proposed in \cite{xu2021rethinking} achieves the state-of-the-art performance on text segmentation with 196K pixel-level labeled training data, which costs hundreds of man-hours but still remains limited in application scenarios (lack of text segmentation data for web text or documentation text). Moreover, existing text segmentation methods only distinguish texts from backgrounds, \textit{i.e.}, instance-level segmentation results are inaccessible to current methods, which usually have limited assistance to downstream tasks.

Recently, generic object segmentation methods like \cite{zhou2018weakly}, which are weakly supervised by categories of the objects in an image rather than pixel-level annotations, have been proposed. However, a few challenging problems exist in text segmentation, \textit{e.g.}, unpredictable fonts, diverse textures, and non-convex contours. Therefore, directly applying weakly-supervised generic segmentation methods to solve the text segmentation task achieves poor performance.

Therefore, our motivation is to propose a text segmentation method that is 1) free from pixel-level annotations, 2) capable of conducting text-instance-wise segmentation. Our key observation is that the rough location of a text instance can be precisely predicted at the decoding stage of an attention-based text recognizer. It sounds reasonable that the text instance segmentation result can thus be obtained by feeding the segmentation head with adequate initial information, \textit{i.e.}, visual features, rough location, and classification of the text instance.

With such prior knowledge as our guideline, we design the proposed method with three modules: text recognition module, text segmentation module, and text adaptive refinement (TAR) module. As a preprocessing step, we employ an off-the-shelf text detector to extract text lines. For each text line image cropped by the detector, the proposed recognition module predicts its text instances and perceives their rough location via the attention mechanism. The insight is to forge the label-free attention maps into segmentation masks by feeding the segmentation module with adequate information: 1) rough attention location as initial seeds, 2) classification results as perceptual knowledge of the text instances 3) visual features as the panorama of the text image. Subsequently, we forge the attention maps again with the proposed TAR module to generate pixel-wise pseudo labels as supervision of the segmentation module. Specifically, TAR is a two-stage refinement module to fit the effective region of each text instance in an iterative manner. Through training, the attention maps shape closer to the text instance, thus further improving the quality of pseudo labels, and it is such iterative mutual enhancement between attention maps and pseudo labels that makes our method work. Additionally, with segmentation masks obtained, it naturally comes to us that mask-augmented contrastive learning between the segmentation results and the input text image can be conducted to enrich the feature representation and suppress the interference caused by complex backgrounds, thus achieving better recognition and segmentation performance. It also benefits existing recognition methods by functioning as a plug-and-play module.

In summary, the contributions of this work are as follows:

\begin{itemize}
    \item This work is the first attempt to propose a weakly-supervised text instance segmentation method. The proposed method couples text segmentation with text recognition, thus bridging the gap between the isolated processes of these two fields.
    \item We propose a text adaptive refinement (TAR) module that provides high-quality pseudo labels through a two-stage iterative procedure with fast inference speed.
    \item We design a mask-augmented contrastive learning strategy to improve the performance of both text recognition and text segmentation. Serving as a plug-and-play solution, it boosts the performance of existing text recognition methods by an average of 2$\%$.
    \item Experimental results show that the proposed method outperforms generic weakly-supervised segmentation methods on text segmentation benchmarks by a considerable margin, specifically by 18.95$\%$ on ICDAR13-FST datasets and 17.80$\%$ on TextSeg datasets.
\end{itemize}

\section{Related Works}
\subsection{Generic Weakly-Supervised Segmentation Methods}
In the last decades, several weakly-supervised methods \cite{zhou2018weakly,ahn2019weakly,cholakkal2019object,kuo2019shapemask,lee2021bbam} for instance segmentation, which usually use the bounding boxes or class activation maps (CAMs) as the rough location for each instance and then refine the rough locations to obtain the fine-grained segmentation masks, are proposed. Although these weakly-supervised generic instance segmentation methods can achieve comparable performance with those methods using pixel-level annotations as supervision, these methods cannot reach satisfying performance when applied to the text segmentation task.
  
\subsection{Text Segmentation Methods}
Traditional methods directly set a threshold for grayscale or employ Markov Random Field (MRF) to separate the foreground and background. In recent years, some methods based on deep neural networks have emerged. In \cite{xu2021rethinking}, the authors put forward TextRnet and a new text segmentation dataset, where TextRnet utilizes the unique text prior such as texture diversity and non-convex contours to achieve state-of-the-art performance on text segmentation benchmarks. The authors of \cite{xu2022bts} mainly focus on Bi-Lingual text segmentation, they propose a Bi-Lingual text dataset along with PGTSNet which contains a plug-in text-highlighting module and a text perceptual module to help distinguish between text languages. Most recently, Textformer\cite{wang2023textformer} leverages the similarities between text components by proposing a multi-level transformer framework to enhance the interaction between text components and image features at different granularities. However, the above-mentioned methods only distinguish texts from the background while the proposed text instance segmentation will be more practical for downstream tasks.

\begin{figure*}[t]
  \centering
  \includegraphics[width=\linewidth]{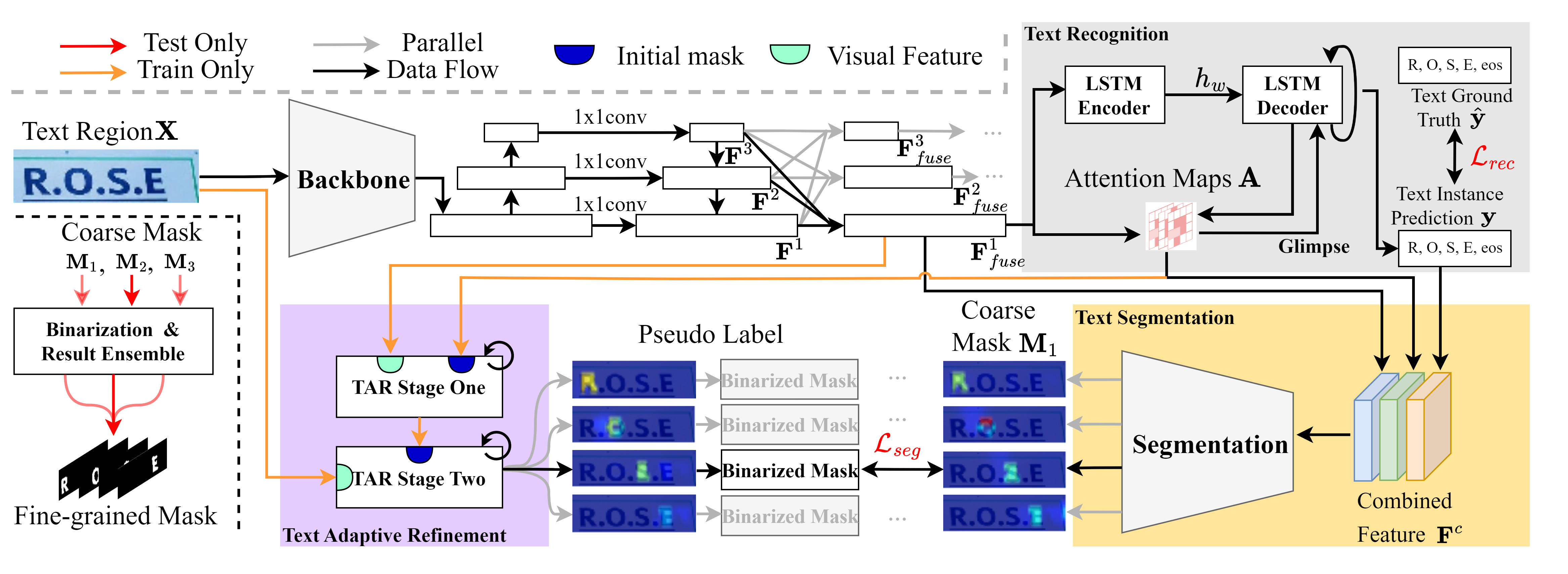}
  \caption{The overall architecture of the proposed method. The recognition module, the segmentation module and the refinement module are on gray, beige and purple backgrounds, respectively. Best viewed in color.}
  \label{fig:architecture}
\end{figure*}



\section{METHODOLOGY}
\textbf{Overview:} As shown in Fig. \ref{fig:architecture}, the proposed method mainly consists of three modules: 1) attention-based text recognition module. 2) text segmentation module. 3) proposed text adaptive refinement module. More details are in the following.

\subsection{Text Recognition Module}
We build our text recognition module following the fashion of SAR \cite{li2019show}, which provides the required attention location while having a light-weighted framework. It is worth noting that the recognition baseline can be replaced by other attention-based recognizers to pursue better performance.


In the wild, texts are varying in scale. To address this challenge, a customized backbone with FPN feature fusion\cite{liu2019learning} is designed to extract features at three levels. Given the input image $\mathbf{X} \in \mathbb{R}^{H \times W \times C}$, the extracted visual features at different levels are denoted as $\mathbf{F}^1 \in \mathbb{R}^{\frac{H}{2} \times \frac{W}{2} \times C} $, $\mathbf{F}^2 \in \mathbb{R}^{\frac{H}{4} \times \frac{W}{4} \times C} $, and $\mathbf{F}^3 \in \mathbb{R}^{\frac{H}{8} \times \frac{W}{4} \times C} $ from bottom to top, respectively. Subsequently, by training adaptive weights at nearly free cost, the fused features manage to keep the scale-invariance between different levels, represented as:
\begin{equation}
  \mathbf{F}^l_{fuse} = \alpha^l \cdot \mathbf{F}^{1 \rightarrow l} + \beta^l \cdot \mathbf{F}^{2 \rightarrow l} + \gamma^l \cdot \mathbf{F}^{3 \rightarrow l}
\end{equation}
Where $l$ denotes the spatial level of fused feature, $\alpha^l$, $\beta^l$, and $\gamma^l$ are trainable adaptive weights. Symbol $\rightarrow$ denotes the resize transformation. As shown in Fig. \ref{fig:architecture}, these fused feature maps with different scales will be parallelly processed and they are uniformly denoted as $\mathbf{F}$ in the following. Subsequently, a 2-layer BiLSTM encoder is applied to extract the holistic feature $\mathbf{h}_w$, which is used to initialize the hidden state of the LSTM decoder. The attention-based decoder aims to model the conditional distribution $P(\mathbf{y}|\mathbf{X})$ in a recurrent manner, which is denoted as follows:
\begin{equation}
  P(\mathbf{y}|\mathbf{X}) = \prod _{t=0}^{T}{P(\mathbf{y}_{t}|\mathbf{g}_{t},\mathbf{h}_{t})}
\end{equation}
where $\mathbf{h}_t$ represents the hidden state of the LSTM decoder. $\mathbf{g}_{t}$ is known as the glimpse vector, which is computed by the following steps:
\begin{equation}
  \mathbf{e}_{i,j} = \tanh(\mathbf{W}_F \mathbf{F}_{ij} + \sum_{p,q \in \mathcal{N}_{ij}} \Tilde{\mathbf{W}}_{p-i,q-j} \cdot \mathbf{F}_{pq} + \mathbf{W}_h \mathbf{h}_t)
\end{equation}
\begin{equation}
  \alpha_{ij} = \text{softmax}(\mathbf{w}_e^\top \cdot  \mathbf{e}_{ij})
\end{equation}
\begin{equation}
  \mathbf{g}_{t} = \sum_{i,j} \alpha_{ij} \mathbf{F}_{ij}, \ \ \ i = 1,...,H; \ j = 1,...,W
\end{equation}
where $\mathbf{F}_{ij}$ is the feature vector at position $(i, j)$ in $\mathbf{F}$, and $\mathcal{N}_{ij}$ is the eight pixels neighboring the position $(i, j)$; $\mathbf{W}_F$, $\mathbf{W}_h$, and $\Tilde{\mathbf{W}}$ are learnable parameters of linear transformations; and $\alpha_{ij}$ is the attention weight at location $(i, j)$. With the glimpse vector $\mathbf{g}_t$ and the hidden state $\mathbf{h}_t$, the $t$-th text instance is predicted by the following transformation:
\begin{equation}
  y_t = \text{softmax}(\mathbf{W}_p[\mathbf{h}_t; \mathbf{g}_t])
\end{equation}
Where $\mathbf{W}_p$ is a linear transformation. Finally, the loss of the text recognition module is as follows:
\begin{equation}
  \mathcal{L}_\text{text} = -\sum_{t={1}}^{T}{\log{P(\hat{y_t}|y_t)}} 
\end{equation}
where $\hat{y}$ is the ground truth of the text in the input image.

\subsection{Segmentation Module}
Through a convolution layer with $1 \times 1$ kernel size, the fused features $\mathbf{F}$, the attention maps $\mathbf{A}$ and the embeddings of the text instance prediction $y$ are concatenated into a combined feature $\textbf{F}^c$, which is taken as the input of the segmentation module. The segmentation module, aiming to generate the coarse mask $\mathbf{M}_i$ for $i$-th text instance, mainly consists of transposed  convolution layers with skip connections. The coarse masks have two channels (\textit{i.e.}, foreground mask and background mask), and are supervised by the binarized pseudo labels generated by the refinement module, which will be detailed in Sec. \ref{refinement}. The loss of the coarse masks is calculated as follows:
\begin{equation}
  \mathcal{L}_\text{seg} = \sum_{t={1}}^{T}{BCE(\mathbf{m}_t, \mathbf{p}_t)} 
\end{equation}
where $\mathbf{p}_t$ denotes the pixel-level pseudo label for the $t$-th coarse mask, and BCE denotes binary cross entropy loss.
\subsection{Refinement Module}
\label{refinement}
\begin{figure}[t]
  \centering
  \includegraphics[width=\linewidth]{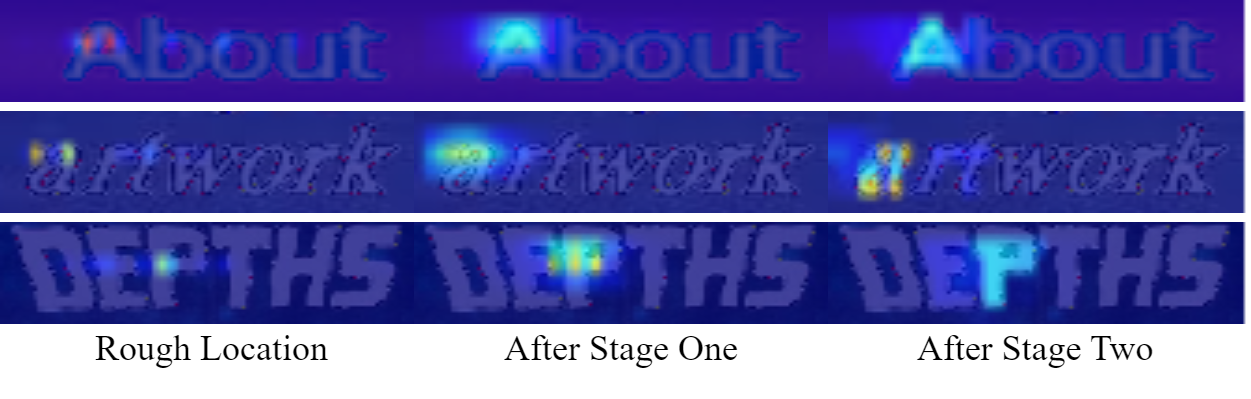}
  \caption{Explanation of the proposed two-stage TAR process.}
  \label{fig:dilation}
\end{figure}

Generic weakly-supervised segmentation method has developed refinement technique \cite{araslanov2020single} based on parameter-free pixel adaptive convolution, which is fast and efficient. On top of that, we propose a two-stage refinement strategy for the text segmentation task. To update each pixel $p^t$ on pseudo label at $t-th$ time step, we have:
\begin{equation}
  p^t = \sum_{a \in \mathcal{A}}{\frac{e^{\overline{k}(\mathcal{V}_p,\mathcal{V}_a)}}{\sum_{n \in \mathcal{A}}{e^{\overline{k}(\mathcal{V}_p,\mathcal{V}_n)}}}} \cdot p^{t-1}
\end{equation}
where $\mathcal{A}$ denotes the collection of all adjacent pixels of $p^t$ in the kernel, and $\mathcal{V}$ denotes the visual feature that actually controls the refinement appearance. k is a kernel function:
\begin{equation}
k(\mathcal{V}_x,\mathcal{V}_y) = \frac{-|\mathcal{V}_x - \mathcal{V}_y|}{\sigma^2_x}
\end{equation}
where $\sigma$ denotes the standard deviation of the
visual intensity inside the kernel. Our key design for text segmentation is that the refinement operation falls into two stages. Please refer to Fig.\ref{fig:dilation} for a vivid explanation: 1) in stage one, we use the backbone feature as visual guidance for TAR, which literally alters the rough location towards the effective receptive field of the text instance, thus making the pseudo label covers more region of the text instance. 2) in stage two, we use the image RGB as the visual guidance for TAR to get the fine-grained pseudo mask of the text instance. It is reasonable that since stage one has covered the major part of the text instance, the text segments of stage two can be more complete.

\subsection{Mask-augmented Contrastive Learning}
With the obtained segmentation results, it naturally comes to us that these segmentation masks can function as positive data augmentation for the model to conduct contrastive learning. Through the recognition baseline, projection features of the input image, the corresponding segmentation mask, the set of the whole training batch, and images other than the input in the batch can be obtained, which is denoted as $\mathbf{P}_i$, $\mathbf{P}_p$, $\mathcal{U}$ and $\mathcal{P}_n$ respectively. We have $l_\text{NCE}$ as the contrastive loss:

\begin{equation}
    l_\text{NCE}(\mathbf{P}_i, \mathbf{P}_p, \mathcal{P}_n) = -log \frac{e^{sim(\mathbf{P}_i, \mathbf{P}_p) / \tau}}{\sum_{\mathbf{P} \in \mathcal{P}_n} e^{sim(\mathbf{P}_i, \mathbf{P}) / \tau}  }
\end{equation}
where $\tau$ is a temperature parameter. $sim(\mathbf{a}, \mathbf{b})$ is to measure the feature distance between $\mathbf{a}$ and $\mathbf{b}$, which is a cosine similarity calculated by
$
    sim(\mathbf{a}, \mathbf{b}) = \frac{\mathbf{a}^\text{T} \mathbf{b}}{ ||\mathbf{a}|| \cdot ||\mathbf{b}|| }
$. Therefore, the loss of the mask-augmented contrastive learning is computed by:
\begin{equation}
    \mathcal{L}_c = \sum_{\mathbf{P}_i\in \mathcal{U}} l_\text{NCE} (\mathbf{P}_i, \mathbf{P}_p, \mathcal{P}_n) + \sum_{\mathbf{P}_i\in \mathcal{U}} l_\text{NCE} (\mathbf{P}_p, \mathbf{P}_i, \mathcal{P}_n)
\end{equation}
Instead of engaging the main recognition-segmentation process, mask-augmented contrastive learning helps to extract a better and more robust presentation that suppresses the influence of complicated backgrounds of text images in the wild.

\subsection{Training Objective} 
The overall loss of the proposed method is calculated as follows:
\begin{equation}
  \mathcal{L} = \mathcal{L}_{seg} + \lambda_{rec} \mathcal{L}_{rec} + \lambda_{c} \mathcal{L}_{c}
\end{equation}
where and $ \lambda_{c}$ are the hyperparameters to balance losses from three modules in our method.

The inference procedure is demonstrated in the bottom-left region of Fig. \ref{fig:architecture}: with coarse masks in all three levels, \textit{i.e.}, $\mathbf{M}_1$, $\mathbf{M}_2$, and $\mathbf{M}_3$ are binarized and ensembled to generate the fine-grained mask as final segmentation result.

\section{Experiments}

\subsection{Datasets and Implementation Details}
\begin{table*}
\centering
\renewcommand\tabcolsep{9.5pt} 
\renewcommand{\arraystretch}{1.0}
  \caption{Performance comparison with other methods, where $Seg^2$ denotes ICDAR2013-FST and TextSeg that with text label, $Seg^5$ denotes the all five text segmentation datasets. TR, UT denotes text recognition datasets and unlabelled text datasets, respectively. $T$,$S$,$B_c$,$B_w$ denote text label, pixel-wise segmentation label, text-instance-wise bounding box annotation, and word-level bounding box annotation, respectively. `pseudo' indicates that the results are tested with binarized pseudo labels.}
  \label{accuracy_public}
  \scalebox{0.85}{
  \begin{tabular}{lccccccc}
    \hline
    Method & Training Data & Supervision & ICDAR13-FST & TextSeg & MLT-S & COCO-TS & Total-Text\\
    \hline
    \multicolumn{8}{l}{\textbf{Fully-Supervised Text Segmentation Methods (Upper-bounds, Not for Comparison)}}\\
    DeeplabV3 \cite{chen2018encoder} & $Seg^5$ & $T+S+B_c$ & 69.27 & 84.07 & 84.63 & 72.07 & 74.44\\
    HRNetV2-W48 \cite{yuan2020object} & $Seg^5$ & $T+S+B_c$ & 70.98 & 85.03 & 83.26 & 68.93 & 75.29\\
    TexRNet \cite{xu2021rethinking} & $Seg^5$ & $T+S+B_c$ & 73.38 & 86.84 & 86.09 & 72.39 & 78.47\\
    \hline
    \multicolumn{8}{l}{\textbf{Weakly-Supervised Instance Segmentation Methods (For Comparison)}}\\
    Zhou \textit{et al.} \cite{zhou2018weakly}& $Seg^2$ & $T$ & 22.34 & 26.79 & - & - & - \\
    Ahn et al. \cite{ahn2019weakly}& $Seg^2$ & $T$ & 35.66 & 38.40 & - & - & - \\
    Cholakkal et al. \cite{cholakkal2019object}&  $Seg^2$ & $T$ & 37.15 & 42.86 & - & - & - \\
    ShapeMask \cite{kuo2019shapemask}& $Seg^2$ & $T+B_c$ & 37.84 & 49.67 & - & - & -\\
    BBAM \cite{lee2021bbam}& $Seg^2$ & $T+B_c$ & 41.60 & 53.59 & - & - & -\\
    \hline
    Ours& $Seg^2$ & $T+B_w$& 48.09 & 59.75 & - & - & - \\
    Ours& $Seg^5$ & $T+B_w$& 51.30 & 64.47 & 62.91 & 55.38 & 56.67 \\
    Ours& $Seg^5$+$TR$& $T+B_w$& 53.80 & 67.99 & 64.75 & 57.23 & 59.85 \\
    \textbf{Ours}& $Seg^5$+$TR$+$UT$& $T+B_w$ & \textbf{60.55} & 71.24 & 72.44 & \textbf{63.14} & \textbf{66.42} \\
    Ours (pseudo)& $Seg^5$+$TR$+$UT$& $T+B_w$ & 60.49 & \textbf{71.39} & \textbf{72.48} & 63.11 & 66.29 \\
  \hline
  \end{tabular}}
\end{table*}

\textbf{Datasets} The training and validation of our network involve text recognition datasets and text segmentation datasets where only text labels are used. To warm up the recognition module, we have several popular datasets from the text recognition field and unlabeled data, details are introduced in \textbf{Appendix}. The joint training stage involves text segmentation datasets, \textit{e.g.}, ICDAR2013-FST \cite{karatzas2013icdar}, TextSeg \cite{xu2021rethinking}, Total-Text \cite{ch2017total}, MLT\_S \cite{bonechi2020weak}, and COCO\_TS \cite{bonechi2019coco_ts}. Different from previous methods, we only utilize text labels instead of pixel-wise labels. The bi-lingual text-related benchmark \cite{chen2021benchmarking} can also be used for training the recognition module.

\begin{table}
\centering

\renewcommand{\arraystretch}{0.9}
\caption{Ablation on different refinement settings.}
  \label{refinement_compare}
  \scalebox{1.0}{
  \begin{tabular}{c|ccc}
    \hline
    Refinement & Iteration & Time (ms) & fIoU (\%) \\
    \hline
    PRMs \cite{zhou2018weakly} & - & 5.6 & 46.21\\
    FC-CRFs \cite{krahenbuhl2011efficient} & - & 89.4 & 64.95\\
    \textbf{TAR (ours)} & \textbf{2 - 8} & \textbf{2.6} & \textbf{71.24}\\
    \hline
\end{tabular}}
\end{table}
\noindent\textbf{Implementation Details} The size of the input is set to 48$\times$160 and the number of channels for the fused feature is set to 512. Please note that three levels of features are parallelly processed by the subsequent modules. The number of iterations for the proposed two-stage TAR module is set to 2 and 8, respectively. Our method is implemented with PyTorch and trained on a single NVIDIA RTX 3090 GPU with 24GB memory. The batch size is set to 32 with a learning rate of 1.0. We conduct warm-up training for the recognition module alone, then jointly train it with the segmentation module. The hyperparameters $\lambda_{rec}$ and $\lambda_{c}$ are empirically set to 1 and 0.1.


\subsection{Comparison with Existing Methods}
We conduct experiments on five aforementioned public text segmentation benchmarks to evaluate our method and listed the performance of the following methods: (a) weakly-supervised generic instance segmentation methods including \cite{zhou2018weakly}, \cite{ahn2019weakly}, \cite{cholakkal2019object}, ShapeMask \cite{kuo2019shapemask}, and BBAM \cite{lee2021bbam}. (b) state-of-the-art text segmentation methods including DeeplabV3 \cite{chen2018encoder}, HRNetV2-W48 \cite{yuan2020object}, and TexRNet \cite{xu2021rethinking}, note that these are fully-supervised methods acting as upper-bound, and our method is not expected to exceed that. Following the fashion of most segmentation manuscripts, we adopt fIoU (foreground Intersection over Union) as the evaluation metric. As shown in Tab. \ref{accuracy_public}, generic weakly-supervised instance segmentation methods with available public sources are retrained with the same training data. Since MLT-S, COCO-TS, and Total-Text do not contain text labels, they are not used to evaluate weakly-supervised methods. Our method, however, is able to train segmentation on the above three datasets by shifting the recognition module to inference mode. Therefore, the proposed method is capable of leveraging text recognition datasets and unlabeled datasets to further enhance performance and suppress overfitting. To sum up, our method significantly outperforms the SOTA weakly-supervised segmentation methods on all available datasets: on ICDAR13-FST, we exceed the former SOTA \cite{lee2021bbam} by 18.95$\%$, while on TextSeg, we take an advantage of 17.80$\%$. Meanwhile, the proposed methods achieve more than 82$\%$ of the fully supervised SOTA's \cite{xu2021rethinking} performance. We also evaluate the performance of the pseudo labels in inference: As shown in the last row of Tab. \ref{accuracy_public}, utilizing the pseudo labels to generate segmentation masks can achieve comparable performance with using the output of the segmentation module.

\subsection{Ablation Studies}
We conduct extensive ablation studies to evaluate the effectiveness of the proposed modules and training strategies. To evaluate the superiority of the proposed TAR, other refinement methods, \textit{i.e.}, FC-CRF \cite{krahenbuhl2011efficient} and PRMs \cite{zhou2018weakly}, which enjoy much popularity among weakly-supervised segmentation methods, are compared. As shown in Tab. \ref{refinement_compare}, the proposed TAR achieves more than 30x speed-up compared to FC-CRFs and 2x speed-up compared to PRMs, and TAR outperforms PRMs and FC-CRFs significantly by 25.03$\%$ and 6.29$\%$ in accuracy.

We also conduct ablation studies to evaluate the effectiveness of mask-augmented contrastive learning, warm-up, and joint training. As shown in Tab. \ref{ablation}, the accuracy of the recognizer in our model will be improved by 1.94\% and fIoU by 0.80\% if our model is equipped with the contrastive learning strategy. Furthermore, the experimental results demonstrate that the warm-up and joint training strategy are necessary to our model, details discussed in \textbf{Appendix}.

\begin{table}
\renewcommand\tabcolsep{7.5pt}
\renewcommand{\arraystretch}{1.0}
  \caption{Ablation studies for contrastive learning, warm-up, and joint training strategies.}
  \label{ablation}
  \scalebox{0.80}{
  \begin{tabular}{ccccc}
    \hline
    Contrastive Learning & Warm-up & Joint Training & Acc &  fIoU \\
    \hline
    & \checkmark & \checkmark  & 91.93 & 70.44\\
    \checkmark & & \checkmark & 79.27 & 52.45 \\
    \checkmark & \checkmark & & 88.04 & 42.38\\
    \checkmark & \checkmark & \checkmark & 93.87 & 71.24\\
    \hline
\end{tabular}}
\end{table}

\begin{table}
\renewcommand\tabcolsep{7.5pt} 
\renewcommand{\arraystretch}{1.0}

  \caption{Plugable mask-augmented contrastive learning.}
  \label{seghelprec}
  \scalebox{0.83}{
  \begin{tabular}{lcc}
    \hline
    Text Recognition Methods & Baseline & Baseline + Segmentation \\
    \hline
    CRNN \cite{shi2016end}&  83.53 & 87.23 (+ \textcolor{green}{3.70})\\
    SAR \cite{li2019show} &  90.85 & 93.10 (+ \textcolor{green}{2.25})\\
    AutoSTR \cite{zhang2020autostr} &  93.88 & 95.06 (+ \textcolor{green}{1.18})\\
    ABINet \cite{fang2021read} &  93.51 & 94.47 (+ \textcolor{green}{0.96})\\
    ours &  91.93 & 93.87 (+ \textcolor{green}{1.94})\\
    \hline
  \label{boost}
\end{tabular}}
\end{table}

\begin{figure}[t]
  \centering
  \includegraphics[width=\linewidth]{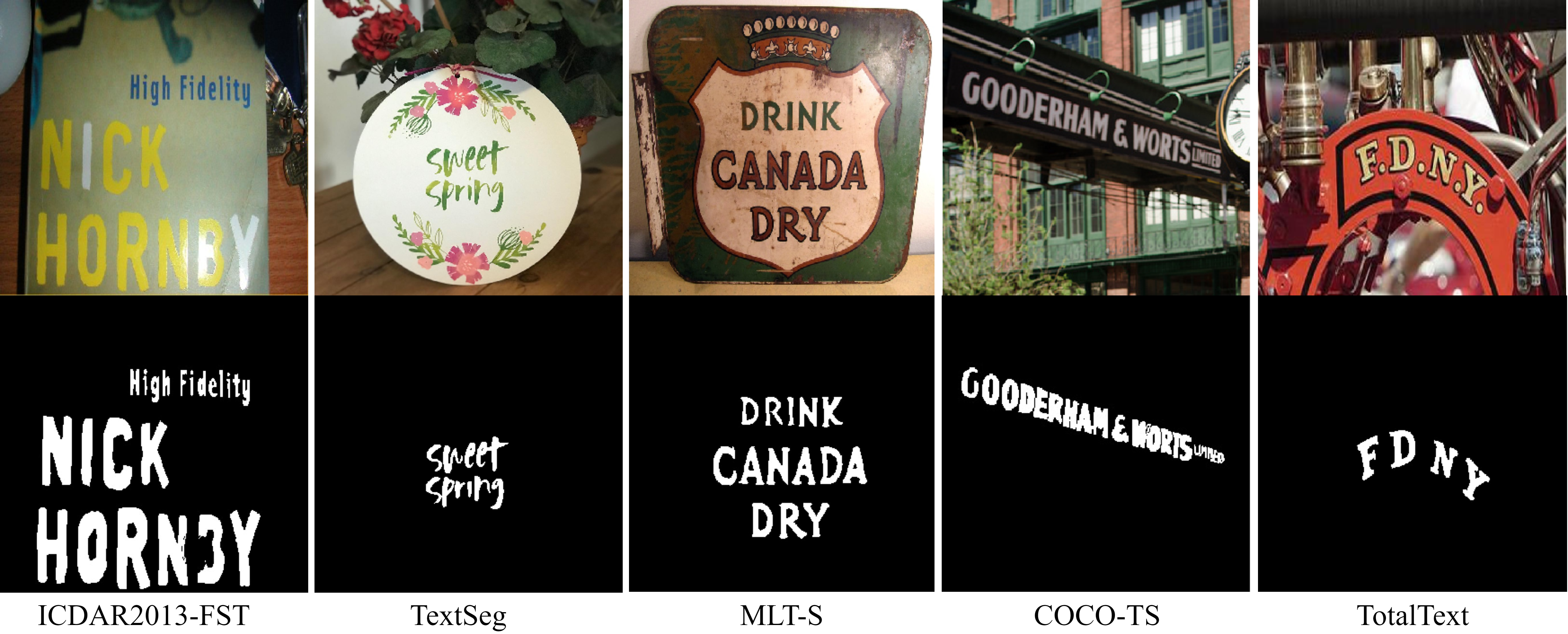}
  \caption{Visualization of our text segmentation results.}
  \label{fig:quality}
\end{figure}

\subsection{Further Discussions and Analysis}
\textbf{Visualization of Segmentation Results}
As shown in Fig. \ref{fig:quality}, there exist several challenges in the five adopted text segmentation datasets: 1) All the selected images are captured in the wild with complex backgrounds. 2) The sample of TextSeg shows challenging aesthetic font types. 3) Unexpected lighting situation occurs in the example of ICDAR2013-FST. In spite of the above three challenges, the visualization results of our method are satisfying.

\textbf{Text Segmentation helps Text Recognition}
Theoretically, the strategy of contrastive learning between original inputs and segmentation masks can be equipped to benefit all existing text recognition methods. We select some of the text recognizers \cite{shi2016end,li2019show,zhang2020autostr,fang2021read} with great popularity in recent years to observe the performance boosts when adopting our strategy. These methods are all fairly re-trained and tested on our datasets (any datasets with text labels). According to the experimental results in Tab. \ref{boost}, all selected methods have achieved a marginal boost in accuracy (2$\%$ in average).

\textbf{Down-stream Applications} Interesting down-stream applications (\textit{e.g.}, image text removal, and text instance style transfer) are demonstrated in \textbf{Appendix}.

\section{Conclusion}
In this paper, we make the first attempt to propose a weakly-supervised text instance segmentation method. Moreover, We propose a two-stage text adaptive refinement module TAR to generate high-quality pseudo labels. In addition, we develop a mask-augmented contrastive learning strategy and make it pluggable, boosting the performance of both our method and existing text recognition methods. The experimental results demonstrate that the proposed method significantly outperforms weakly-supervised segmentation methods on ICDAR13-FST by 18.95$\%$ and on TextSeg by 17.80$\%$, and it achieves 82\% of the fully supervised SOTA’s performance. Last but not least, the proposed method drives multiple downstream applications in practical scenarios.
{\small
\bibliographystyle{IEEEbib}
\bibliography{icme2022template}
}
\appendix
\noindent \Large \textbf{Appendix}
\normalsize
\section{Down-stream Applications}
Compared with \cite{xu2021rethinking}, the proposed method, which can produce the text instance segmentation mask, provides more powerful and controllable operations for downstream applications.

\begin{figure}[ht]
  \centering
  \includegraphics[width=\linewidth]{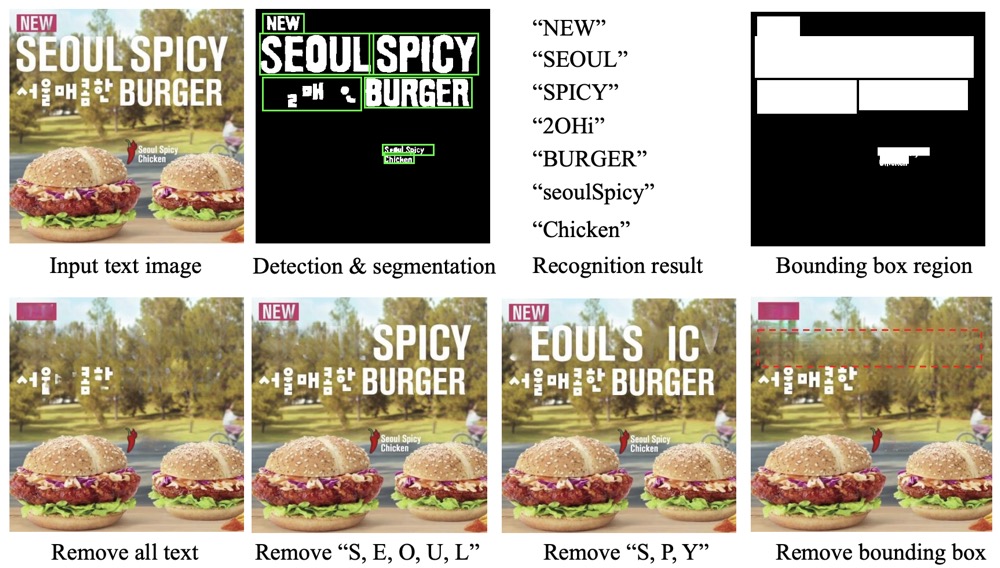}
  \caption{An example of text instance removal.}
  \label{removal}
\end{figure}

\begin{table*}[ht]
\centering

\renewcommand{\arraystretch}{1.0}
\caption{Runtime of all methods}
  \label{runtime}
  \scalebox{1.0}{
  \begin{tabular}{|c|c|c|c|c|c|c|}
    \hline
    Methods & Zhou et al. \cite{zhou2018weakly} & Ahn et al. \cite{ahn2019weakly} & Cholakkal et al. \cite{cholakkal2019object}  & ShapeMask \cite{kuo2019shapemask} & BBAM \cite{lee2021bbam} & Ours\\
    \hline
    Runtime(ms) & 18.9 & 30.4 & 15.9  & 89.5 & 48.0 & 28.7\\
    \hline
\end{tabular}}
\end{table*}

\begin{table*}[ht]
\centering

\renewcommand{\arraystretch}{1.0}
\caption{Adjusting of hyper-parameters $\lambda_{rec}$ and $\lambda_{c}$, the f-IoU performance is tested on TextSeg datasets.}
  \label{param}
  \scalebox{1.0}{
  \begin{tabular}{|c|c|c|c|c|c|c|c|c|}
    \hline
    \multirow{2}*{Params.}& $\lambda_{rec}$ $|$ $\lambda_{c}$  & $\lambda_{rec}$ $|$ $\lambda_{c}$ & $\lambda_{rec}$ $|$ $\lambda_{c}$  & $\lambda_{rec}$ $|$ $\lambda_{c}$  & $\lambda_{rec}$  $|$ $\lambda_{c}$ & $\lambda_{rec}$  $|$ $\lambda_{c}$& $\lambda_{rec}$  $|$ $\lambda_{c}$& $\lambda_{rec}$  $|$ $\lambda_{c}$ \\
    \cmidrule(r){2-9}
    ~ &  0.1 $|$  0.1 &  0.1 $|$ 1 &  1 $|$ 0.1  &   1 $|$  1 &  1 $|$  10 &  10 $|$ 10 &  0.01 $|$ 1 &  1 $|$ 100\\
    \hline
    f-IoU & 69.85 & 70.66 & \textbf{71.24} & 71.18 & 70.43 & 70.57 & 70.89& 69.01\\
    \hline
\end{tabular}}
\end{table*}

\textbf{Text Instance Removal} Scene text removal is urgently needed by some industries \textit{e.g.}, advertising, film, television, etc. In these scenarios, it is a challenging task to remove unwanted text in videos or images while keeping other contents unchanged. Although \cite{xu2021rethinking} achieves seemingly satisfactory text removal results, its application scenarios are limited due to its semantic segmentation nature, \textit{i.e.}, this method erases every text in the image, wanted or not. Different from previous methods, our method enables the user to remove specific unwanted text instances while leaving others untouched. The visualization of this application is demonstrated in Fig. \ref{removal}.

\begin{figure}[ht]
  \includegraphics[width=\linewidth]{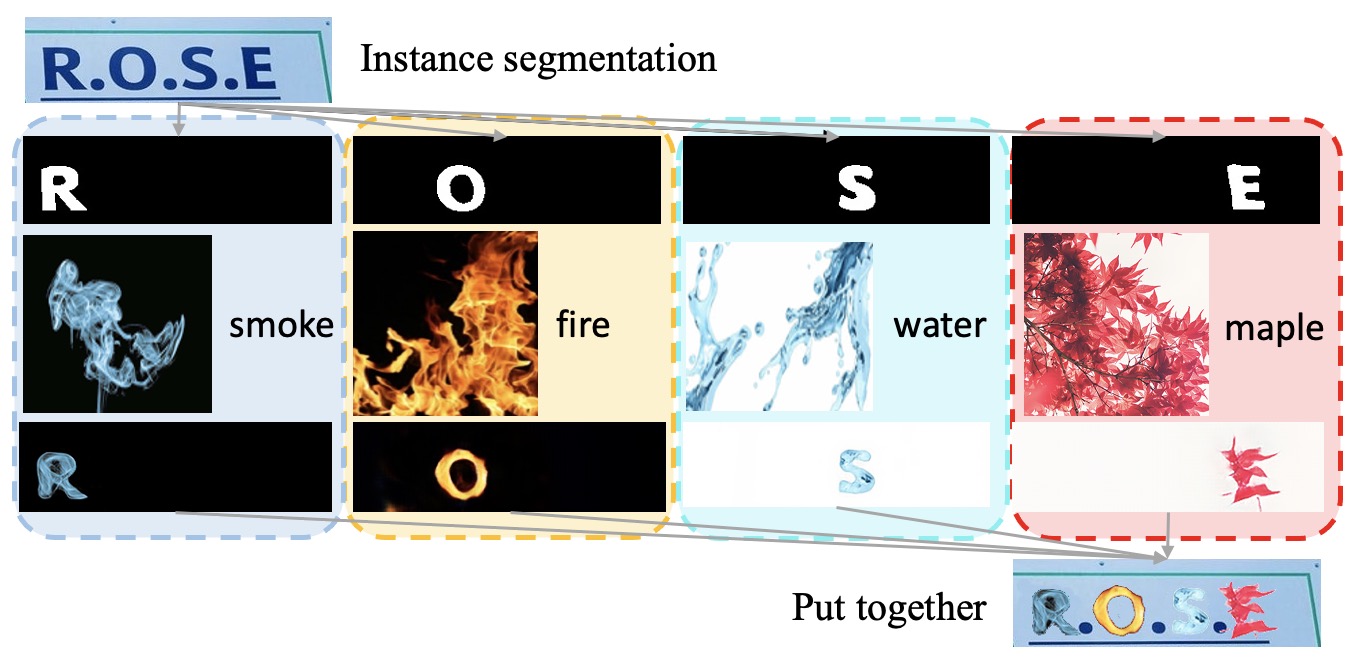}
  \caption{An example of text instance style transfer.}
  \label{transfer}
\end{figure}

\textbf{Text Instance Style Transfer} Text segmentation can help transfer a font style to a given text image, which usually requires accurate text masks. In this work, we upgrade font style transfer to text-instance-level, which means that the user can transfer different styles to different text instances. An example of the text-instance-wise style transfer is illustrated in Fig. \ref{transfer}. By combining the controllable style transfer and text-instance removal, one can easily conduct data augmentation on text images in the wild to produce high-value synthetic datasets with complex real-world background for text-related researches.

\begin{figure}[ht]
  \centering
  \includegraphics[width=\linewidth]{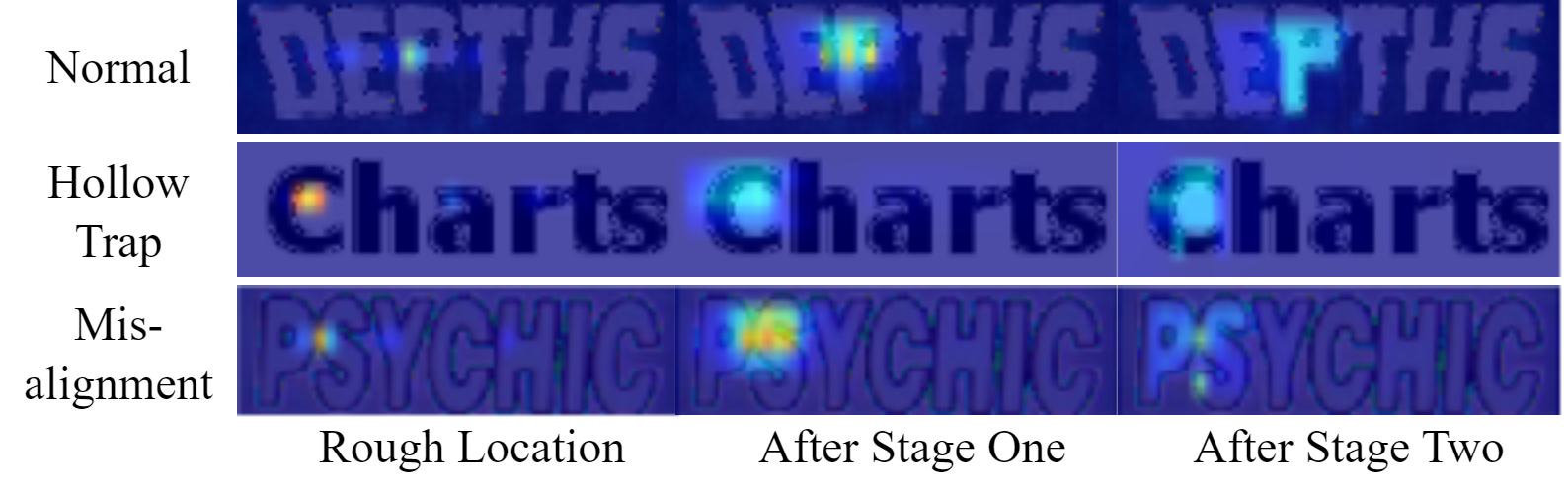}
  \caption{Case study of pseudo labels generated by rough attention location with different qualities. Rows titled `Normal' and `Hollow Trap' show normal cases, `Misalignment' denotes the failure case. Note that the visualization results are captured at the beginning of the joint training stage.}
  \label{warm-up}
\end{figure}

\section{Warm-Up and Joint Training}
\textbf{Importance of Warm-Up for Recognizer}
\label{discussion}
As mentioned above, the proposed method should warm up the recognizer in advance. Since pseudo labels for the segmentation module derive from the attention maps generated by the recognizer, the refinement module will not produce effective pseudo labels if the recognizer has a wrong awareness of text instance position, as vividly shown in Fig. \ref{warm-up}. The bottom row shows that the refinement module produces imprecise pseudo labels based on misaligned attention maps, which will further mislead the segmentation module. Therefore, the warm-up strategy is essential for the proposed method to generate accurate attention localization. Additionally, as in the middle row, an interesting observation is that for text instances like `C`, 'O', and 'D', the refinement of their initial attention sometimes gets misled by the hollow structures of their own (inevitable), and it will take a long period of joint training for the model to manage escaping from such hollow traps.

\begin{figure}
  \centering
  \includegraphics[width=\linewidth]{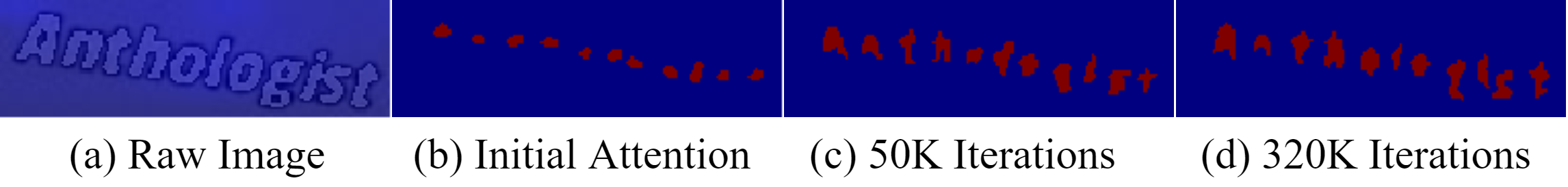}
  \caption{The visualization of attention location during joint training. The attention is binarized for better understanding.}
  \label{joint-training}
\end{figure}

\textbf{Necessity of Joint Training}
After warming up the recognition module, the segmentation module will be jointly trained to achieve better performance. At the warm-up stage, the recognizer can generate a rough location for each text instance (Fig. \ref{joint-training}(b)). However, unlike CAMs \cite{zhou2016learning}, the attention mechanisms for text recognizers tend to focus on the center of text instance. Refining such focused attention as initial seeds cannot completely perceive the instance boundary. Therefore, when the recognition module and segmentation module are jointly trained, the attention maps generated by the recognizer gradually learn to be closer to the shape of the corresponding text instance (shown in Fig. \ref{joint-training}(c) and Fig. \ref{joint-training}(d)), which results in pseudo labels with better quality for the segmentation module. In short, the effectiveness of our method is based on such mutual enhancement between attention maps and pseudo labels. Additionally, the ablation experiments also validate the necessity of joint training. It also needs to be mentioned that we observed the recognition loss fluctuates greatly at the beginning of the joint training. The phenomenon, most likely caused by the reshaping of attention maps, will vanish with longer training.

\textbf{Text Recognition Data for Warm-up Training}
For the warm-up of text recognition datasets, we use Syn90K \cite{jaderberg2014synthetic}, SynthText \cite{gupta2016synthetic}, IC13 \cite{karatzas2013icdar}, etc, which follows the fashion of recent text recognition methods. In addition, unlabelled datasets, \textit{i.e.}, BOOK32 \cite{iwana2016judging}, TextVQA \cite{singh2019towards} and ST-VQA \cite{biten2019scene}, are used through a semi-supervised training strategy. To enable semi-supervised training, all we need is to switch the text recognition module in evaluation mode.

\section{Additional Experiments}
We put the experiment results of the empirical configuration of weighting balance and inference time in Tab. \ref{param} and Tab. \ref{runtime}.
\end{document}